\documentclass[letterpaper, 10 pt, conference]{ieeeconf}  %
\usepackage{url}
\usepackage{cite}
\usepackage{balance}
\usepackage{graphicx}
\usepackage{amsfonts}
\usepackage{fancyhdr}
\usepackage{comment}
\usepackage{times}
\usepackage{amsmath}
\usepackage{changepage}
\usepackage{amssymb}
\usepackage{enumerate}
\usepackage{algorithm}
\usepackage{algorithmic}
\usepackage{bm}
\usepackage{subfigure}
\usepackage{calligra}
\usepackage{multirow}
\usepackage{booktabs}
\usepackage{mathtools}
\usepackage{arydshln}
\usepackage{latexsym}
\usepackage{amsmath}
\usepackage{amssymb}

\IEEEoverridecommandlockouts                              %
\newcommand\numberthis{\addtocounter{equation}{1}\tag{\theequation}}

\title{\LARGE \bf
	Automatic Calibration of Multiple 3D LiDARs in Urban Environments
}

\author{Jianhao Jiao$^{1}$, Yang Yu$^{1}$, Qinghai Liao$^{1}$, Haoyang Ye$^{1}$, Ming Liu$^{1}$
\thanks{$^{1}$
	Jianhao Jiao, Yang Yu, Qinghai Liao, Haoyang Ye and Ming Liu are with the Robotics and Multi-Perception Laborotary, Robotics Institute, The Hong Kong University of Science and Technology, Hong Kong SAR, China. {\tt\small \{jjiao, yang.yu, qinghai.liao, hy.ye, eelium\}}@ust.hk }
}

\def\hlinew#1{%
	\noalign{\ifnum0=`}\fi\hrule \@height #1 \futurelet
	\reserved@a\@xhline}

\begin{document}
\IEEEpeerreviewmaketitle
\maketitle
\thispagestyle{empty}
\pagestyle{empty}

\begin{abstract}
Multiple LiDARs have progressively emerged on autonomous vehicles for rendering a wide field of view and dense measurements.
However, the lack of precise calibration negatively affects their potential applications in localization and perception systems.
In this paper, we propose a novel system that enables automatic multi-LiDAR calibration without any calibration target, prior environmental information, and initial values of the extrinsic parameters.
Our approach starts with a hand-eye calibration for automatic initialization by aligning the estimated motions of each sensor. 
The resulting parameters are then refined with an appearance-based method by minimizing a cost function constructed from point-plane correspondences. 
Experimental results on simulated and real-world data sets demonstrate the reliability and accuracy of our calibration approach.
The proposed approach can calibrate a multi-LiDAR system with the rotation and translation errors less than $\textbf{0.04}$ [rad] and $\textbf{0.1}$ [m] respectively for a mobile platform.
\end{abstract}

\section{Introduction}
\label{sec.introduction}

\subsection{Motivation}
\begin{figure}
	\centering
	\includegraphics[width=0.4\textwidth]{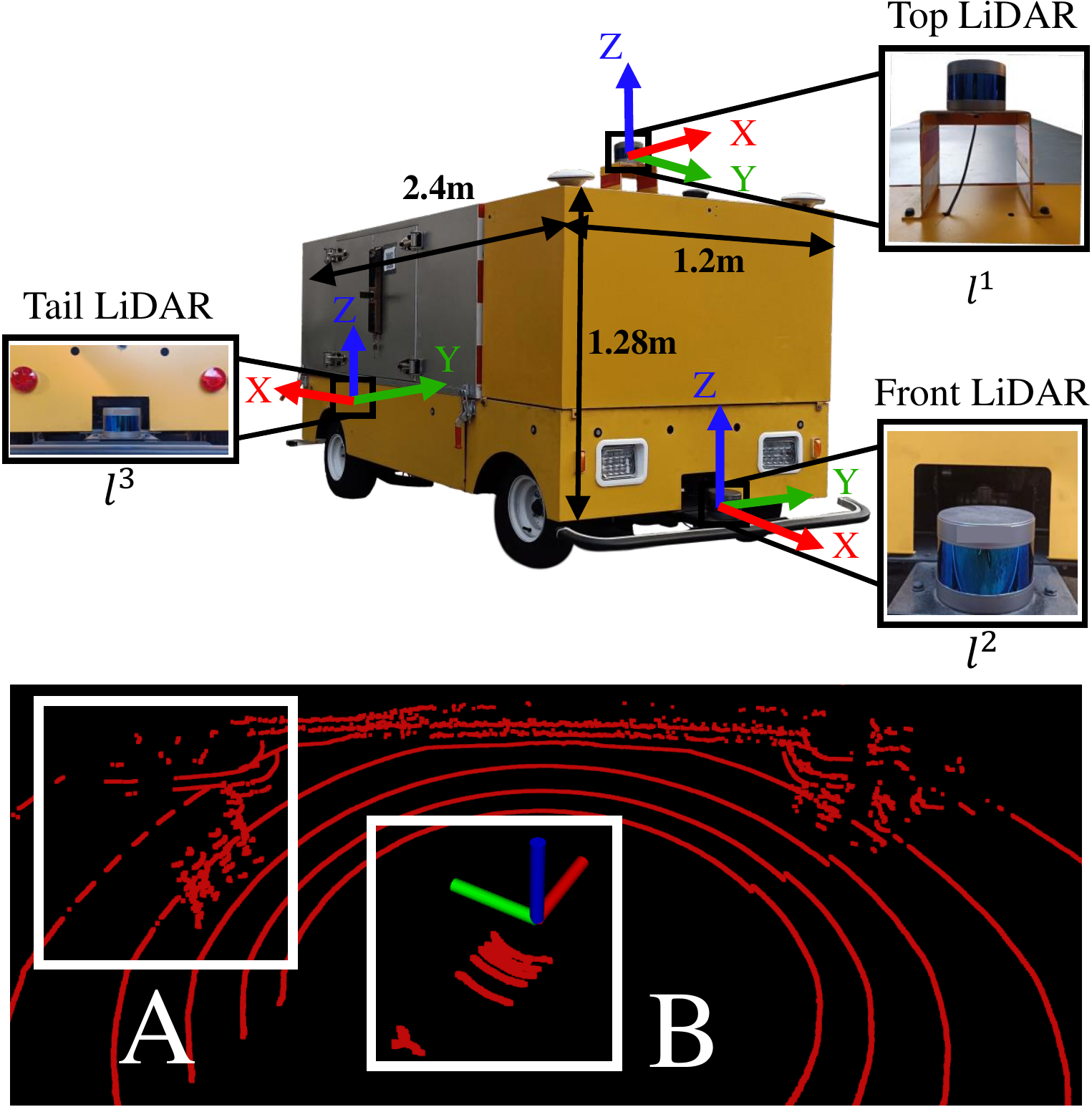}
	\caption{(Top) Our mobile platform consists of a multi-LiDAR system with unknown extrinsic parameters. (Bottom) A point cloud captured by $l_{1}$. The white boxes indicate two drawbacks presented in a single LiDAR configuration: (A) measurement sparsity and (B) occlusion.}    
	\label{fig.platform} 	
\end{figure}

Accurate extrinsic calibration has become increasingly essential for the broad applications of multiple sensors. Numerous research work has been studied on  \cite{heng2013camodocal,taylor2016motion,zhou2018automatic}.
Over the past decade,  Light Detection and Ranging (LiDAR) sensors have appeared as a dominant sensor in mobile robotics for their active nature of providing accurate and stable distance measurements. They have been widely applied in 3D reconstruction \cite{behley2018efficient}, localization \cite{shan2018lego}, and object detection \cite{du2018general}. 
However, the common drawbacks of LiDARs are the low spatial resolution on measurements and the sensibility to occlusion.
These limit its potential utilization in robotic systems. Fig. \ref{fig.platform} (bottom) presents two examples, which shows a point cloud captured by the top LiDAR. In the block A, the pedestrians and vehicles are scanned with a few points, making the detection of these objects challenging. About the block B, points are gathering together since they are occluded by the vehicle's body.
To solve these problems, the development of a multi-LiDAR configuration is necessary.

Traditional calibration techniques are realized by either placing markers in scenes or hand-labeled correspondences. But these approaches suffer from impracticality and limited scalability to the multi-LiDAR configuration. 
There are surprisingly few discussions on calibrating multiple 3D LiDARs. Moreover, the majority of current approaches involve one or more of the following assumptions: prior knowledge about the structure of environments \cite{xie2018infrastructure}, usage of additional sensors \cite{gao2010line}, and user-provided initial values of the extrinsic parameters \cite{choi2016extrinsic}. 
Inspired by the progress on the hand-eye calibration, we find that the extrinsic parameters can be directly recovered from individual motions provided by each LiDAR. 
In addition, the geometric features in environments also form constraints to solve the calibration.
Therefore, we conclude that the complementary usage of these approaches is a prospective solution to the multi-LiDAR calibration.

\begin{figure*}[t]
	\vspace{0.2cm}
	\centering
	\includegraphics[width=0.83\textwidth]{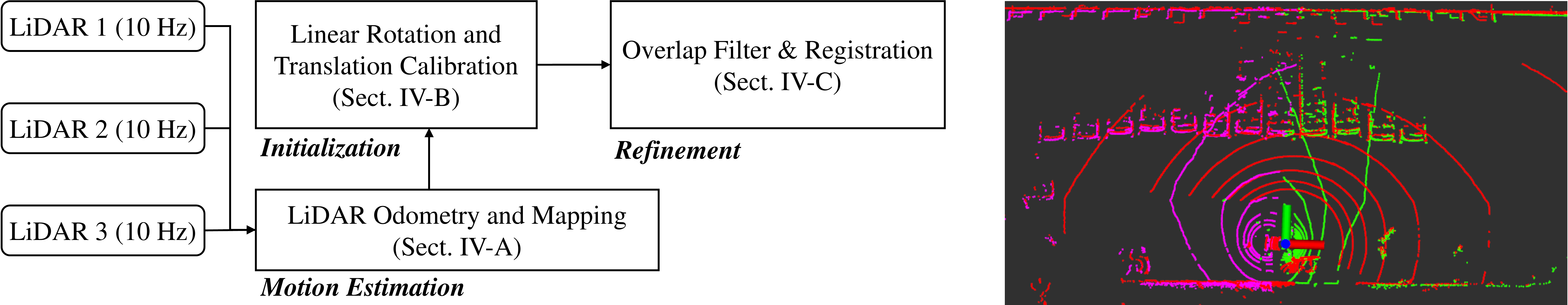}
	\caption{This figure illustrates the full pipeline of the proposed approach (left) and the fused point clouds after calibration (right). Note that the red, green, and purple point clouds are captured by the top LiDAR, front LiDAR, and tail LiDAR respectively.}    
	\label{fig.pipeline} 	
	\vspace{-0.4cm}
\end{figure*}  	

\subsection{Contribution} 
In this paper, we proposed a novel system which allows automatic calibration of multiple LiDARs in urban environments.
This system consists of three components: $\textbf{motion estimation}$ of each sensor, motion-based $\textbf{initialization}$ , and appearance-based $\textbf{refinement}$ .
We show a variety of experiments to demonstrate the reliability and accuracy of our proposed approach.
The contributions of this paper are summarized as following:
\begin{itemize}
	\item 
	A pipeline to automatically calibrate the extrinsic parameters of multiple LiDARs which releases the assumptions of calibration targets, prior knowledge about surroundings, and initial values given by users.

	\item A complementary usage of motion-based method for initialization and appearance-based method for refining the estimates.
	\item Extensive evaluation experiments on simulated and real-world data sets.
\end{itemize}

\subsection{Organization}
The rest of the paper is organized into the following sections. In Sect. \ref{sec.related_work}, the relevant literature is discussed. An overview of the complete system pipeline is given in Sect. \ref{sec.overview}. The methodology of our approach which includes motion estimation, automatic initialization, and refinement is introduced in Sect. \ref{sec.methodology}, followed by experimental results presented in Sect. \ref{sec.experiment}. Finally, Sect. \ref{sec.conclusion} summarizes the paper and discusses possible future work.

\section{RELATED WORK}
\label{sec.related_work}

Besides the multi-LiDAR calibration, there are extensive discussions related to our work on the calibration among LiDARs, cameras, and IMUs. In this section, we  categorize them as appearance-based or motion-based methods.

\subsection{Appearance-based approaches}
Appearance-based approaches that recover the spatial offset using appearance cues in surroundings are considered as a category of registration problem. The key challenge is searching correspondences among data.
Artificial markers that are observable to sensors have been prevalently used to acquire correspondences. 
Gao et al. \cite{gao2010line} proposed a multi-LiDAR calibration method using point constraints from the retro-reflective targets placed in scenes. 
Choi et al. \cite{choi2016extrinsic} determined spatial offset of dual 2D LiDARs by employing the appearance of two orthogonal planes.
For calibrating a LiDAR with a camera. 
Zhou et al. \cite{zhou2018automatic} demonstrated a technique to form line and plane constraints between the two sensors in the presence of a chessboard, while Liao et al. \cite{qinghai2019extrinsic} published a toolkit using an arbitrary polygon, which is more general than the previous approach. 
All these methods require fixed markers, which are time-consuming and labor intensive. Our approach only depends on the common features such as edges and planes, which is more general and effective.

The automatic markerless calibration in an arbitrary scene has led the trend recently. He
et al. \cite{he2013pairwise} extracted geometric features among scan points to achieve robust registration, and their work was extended to a challenging scenario \cite{he2014calibration}.
Levinson \cite{levinson2013automatic} first put forward an online calibration of a camera-LiDAR system. This is accomplished by aligning edge points with image contours and minimizing a cost function. Other metrics for matching sensor data are proposed including Renyi Quadratic Entropy \cite{maddern2012lost}, Mutual Information \cite{pandey2012automatic}, and Gradient
Orientation Measure \cite{taylor2012mutual}. However, the success of these approaches highly relies on the prior knowledge of initial values. Compared with their methods, our approach can recover the initial extrinsic parameters from sensor's motions, which enables calibration in poor human intervention.

\subsection{Motion-based approaches}
The motion-based approaches treat calibration as a well-researched hand-eye calibration problem \cite{horaud1995hand}, where the extrinsic parameters are computed by combining the motions of all available sensors. The hand-eye calibration problem is usually referred to solve $\mathbf{X}$ in $\mathbf{AX = XB}$, where $\mathbf{A}$ and $\mathbf{B}$ are the motions the two sensors undergo, and $\mathbf{X}$ is the transformation between them. As described in \cite{daniilidis1999hand, kabsch1978discussion}, this problem has been addressed since 1980s. 
The ongoing research has extended this motion-based 
approaches to calibrate multiple sensors in outdoor environments. Heng et al. \cite{heng2013camodocal} proposed CamOdoCal, a versatile algorithm with a bundle adjustment to calibrate four cameras. For more general sensor configurations, Taylor et al. \cite{taylor2015motion} provided a solution to calibrate sensors in three different modes. As presented in \cite{taylor2016motion,qin2018online}, these approaches can also be utilized to estimate temporal offset between sensors. Additionally, several state-of-the-art visual-inertial navigation systems adopted the motion-based approaches to achieve online calibration of camera-IMU transformation \cite{qin2018vins}. 
Although the motion-based calibration has been extensively developed, 
the accuracy of the results is easily affected by the accumulated drifts of the estimated motions.
In contrast, our method takes advantages of extracted appearance features to refine the motion-based calibration of multiple LiDARs.

\section{Overview}
\label{sec.overview}

The notations are defined as following. 
We denote $[0, K]$ the time interval during calibration, and define $\{l_{k}^{i}\}$ as the sensor coordinate system of the $i^{th}$ LiDAR $l^i$ at timestamp $k\in[0, K]$. 
The $x\textendash$, $y\textendash$ and $z\textendash$ axes of these coordinate systems are pointing forward, left and upward respectively. 
We denote $I$ the number of LiDARs to be calibrated, and $l^1$ the reference LiDAR among them.
The transformation, rotation, and translation from $\{a\}$ to $\{b\}$ are denoted by $\mathbf{T}_{b}^{a}$, $\mathbf{R}_{b}^{a}$, and $\mathbf{t}_{b}^{a}$ respectively. The unknown transformations from $\{l^1\}$ to $\{l^i\}$ are thus represented as $\mathbf{T}_{l^{i}}^{l^{1}}$ or $(\mathbf{R}_{l^{i}}^{l^{1}},\mathbf{t}_{l^{i}}^{l^{1}})$.
Additionally, the point cloud perceived by $l^i$ at $k$ is denoted by $\mathcal{P}^{l^{i}_{k}}$, and the extracted ground points are denoted by $\mathcal{G}^{l^{i}_{k}}$. 
We assume that LiDARs are synchronized, where point clouds are captured at the same time. We also assume that the vehicle is able to perform sufficient motions over a planar surface. 
With the pre-defined notations and assumptions, the calibration problem can be defined as:

\textbf{\textit{Problem:}} Given a sequence of point clouds $(\mathcal{P}^{l^{i}_{k}}, i\leqslant I, k\leqslant K])$ during the calibration, compute the extrinsic parameters of a multi-LiDAR system 
by combining the estimated motions with surrounding appearance.

The pipeline of our proposed calibration system consists of three phases, as illustrated in Fig. \ref{fig.pipeline}. The first phase takes point clouds as input, and results in the incremental motions of each LiDAR at within a time interval $[k-1, k]$ (Sect. \ref{sec.methodology_motion_estimation}). The second phase initializes $\mathbf{T}_{l^{i}}^{l^{1}}$ using a least-squares solution (Sect. \ref{sec.methodology_initialization}). Finally, the third phase utilizes the appearance cues in surroundings to register difference LiDARs for refinement (Sect. \ref{sec.methodology_refinement}).

\section{Methodology}
\label{sec.methodology}

\begin{figure}[]
	\vspace{0.1cm}
	\centering
	\includegraphics[width=0.35\textwidth]{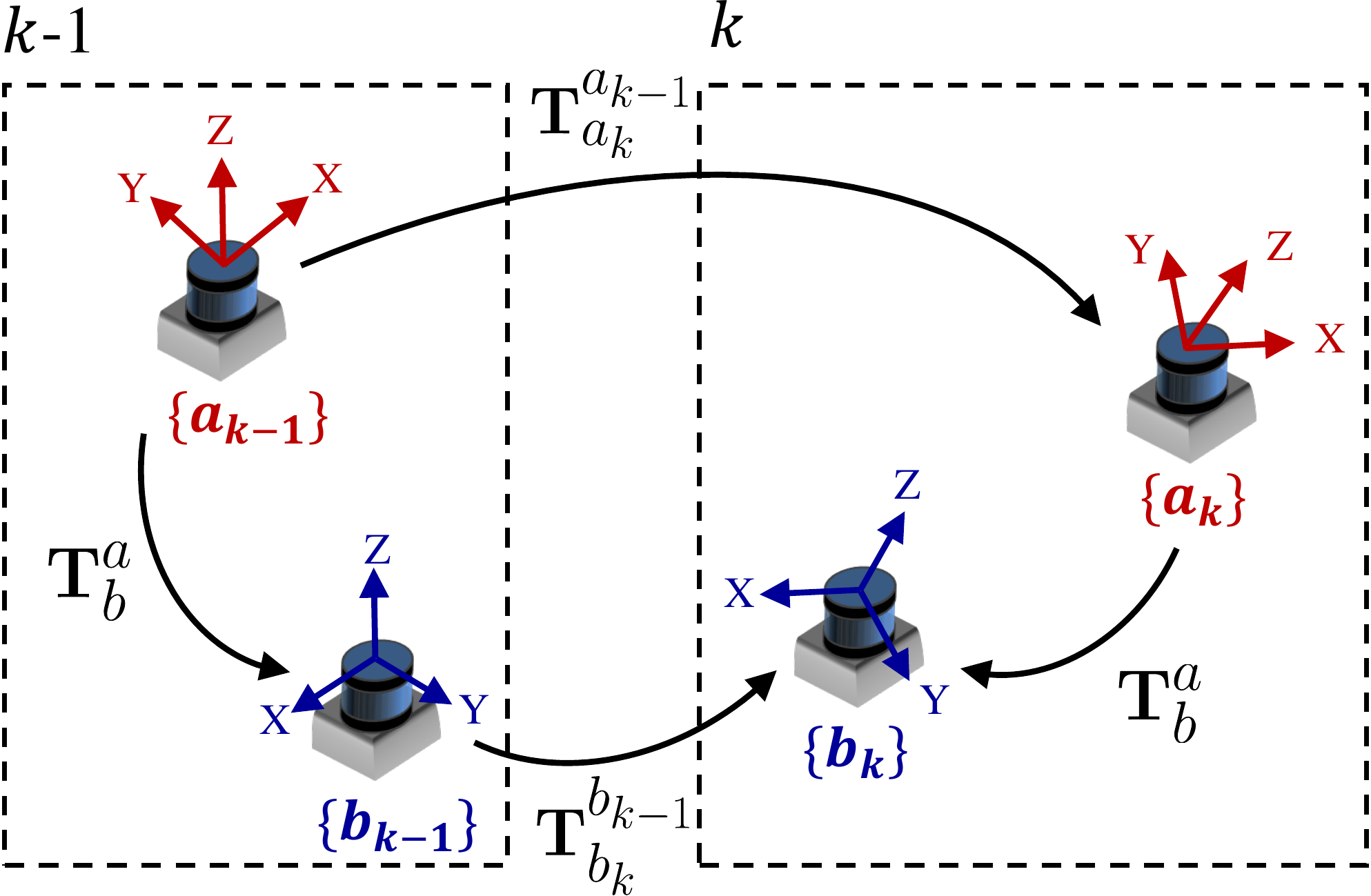}
	\caption{The transformations between different LiDARs at $[k-1, k]$.}    
	\label{fig.hand_eye_calibration} 
\end{figure}  	

\subsection{Motion Estimation}
\label{sec.methodology_motion_estimation}
	To calculate a set of incremental motions between consecutive frames of each LiDARs, the LeGO-LOAM algorithm \cite{shan2018lego} is used. This method makes use of geometric features in environments to estimate the ego-motion. In our implementation, the individual transformations of all the sensors and the extracted ground point clouds are further used to provide constraints to the extrinsic parameters, which will be discussed in Sect. \ref{sec.methodology_initialization} and Sect. \ref{sec.methodology_refinement}.

\subsection{Initialization}
\label{sec.methodology_initialization}

	With $l^{1}$ as the reference, we present a method of calibrating $l^{i}$ with $l^{1}$ pairwise. To simplify the notations, we replace $\{l^{1}\}, \{l^{i}\}$ with $\{a\}, \{b\}$ to indicate the coordinate system of reference sensor and target sensor respectively. The constant transformation of two LiDARs can be initialized by aligning their estimated motions. 
	Fig. \ref{fig.hand_eye_calibration} depicts the relationship between the motions of two LiDARs and their relative transformation. As the vehicle moves, the extrinsic parameters can be recovered using these motions for any $k$:
	\begin{align}
		\mathbf{T}_{a_{k}}^{a_{k-1}}\mathbf{T}_{b}^{a} 
		&= 
		\mathbf{T}_{b}^{a}\mathbf{T}_{b_{k}}^{b_{k-1}},
		\label{equ.hand_eye_T}
	\end{align}	
	where \eqref{equ.hand_eye_T} can be decomposed in terms of its rotation and translation components with the following two equations:
	\begin{align} 
	\mathbf{R}_{a_{k}}^{a_{k-1}}\mathbf{R}_{b}^{a} 
	&= 
	\mathbf{R}_{b}^{a}\mathbf{R}_{b_{k}}^{b_{k-1}}, 	
	\label{equ.hand_eye_calibration_1} 
	\\
	(\mathbf{R}_{a_{k}}^{a_{k-1}}-\mathbf{I}_{3})\mathbf{t}_{b}^{a} 
	&= 
	\mathbf{R}_{b}^{a}\mathbf{t}_{b_{k}}^{b_{k-1}} - \mathbf{t}_{a_{k}}^{a_{k-1}}.
	\label{equ.hand_eye_calibration_2} 
	\end{align}
	
	The method described in \cite{heng2013camodocal} is used to solve these two equations. Based on \eqref{equ.hand_eye_calibration_1}, the pitch-roll rotation can be calculated directly using the estimated rotations, while the yaw rotation, the translation can be computed using \eqref{equ.hand_eye_calibration_2}. 
	
	\subsubsection{Outlier Filter}
		The motions of dual sensors have two constraints that are independent from the extrinsic parameters, which were proposed as the screw motion in \cite{chen1991screw}:
		\begin{align}
			\theta_{a_{k}}^{a_{k-1}} 
			&=
			\theta_{b_{k}}^{b_{k-1}}
			\label{equ.screa_motion_rotation}
			\\
			\mathbf{r}_{a_{k}}^{a_{k-1}} \cdot			
			\mathbf{t}_{a_{k}}^{a_{k-1}} 
			&=
			\mathbf{r}_{b_{k}}^{b_{k-1}} \cdot
			\mathbf{t}_{b_{k}}^{b_{k-1}},
			\label{equ.screa_motion_translation}			
		\end{align}
		where $\theta$ denotes the angle of a rotation matrix $\mathbf{R}$, and $\mathbf{r}$ is the corresponding rotation axis\footnote{The rotation angle and axis are calculated using the $\log(\cdot)^{\vee}$ operator such that $\bm{\phi}=\log(\mathbf{R})^{\vee}, \bm{\phi} = \theta\mathbf{r}$.}. 
		The screw motion residuals include rotation and translation residuals, which are calculated as: $|\theta_a - \theta_b|$,  $\|\mathbf{r}_{a}\cdot\mathbf{t}_{a}-\mathbf{r}_{b}\cdot\mathbf{t}_{b}\|^{2}$.
			
\begin{figure}[]
			\centering
			\includegraphics[width=0.48\textwidth]{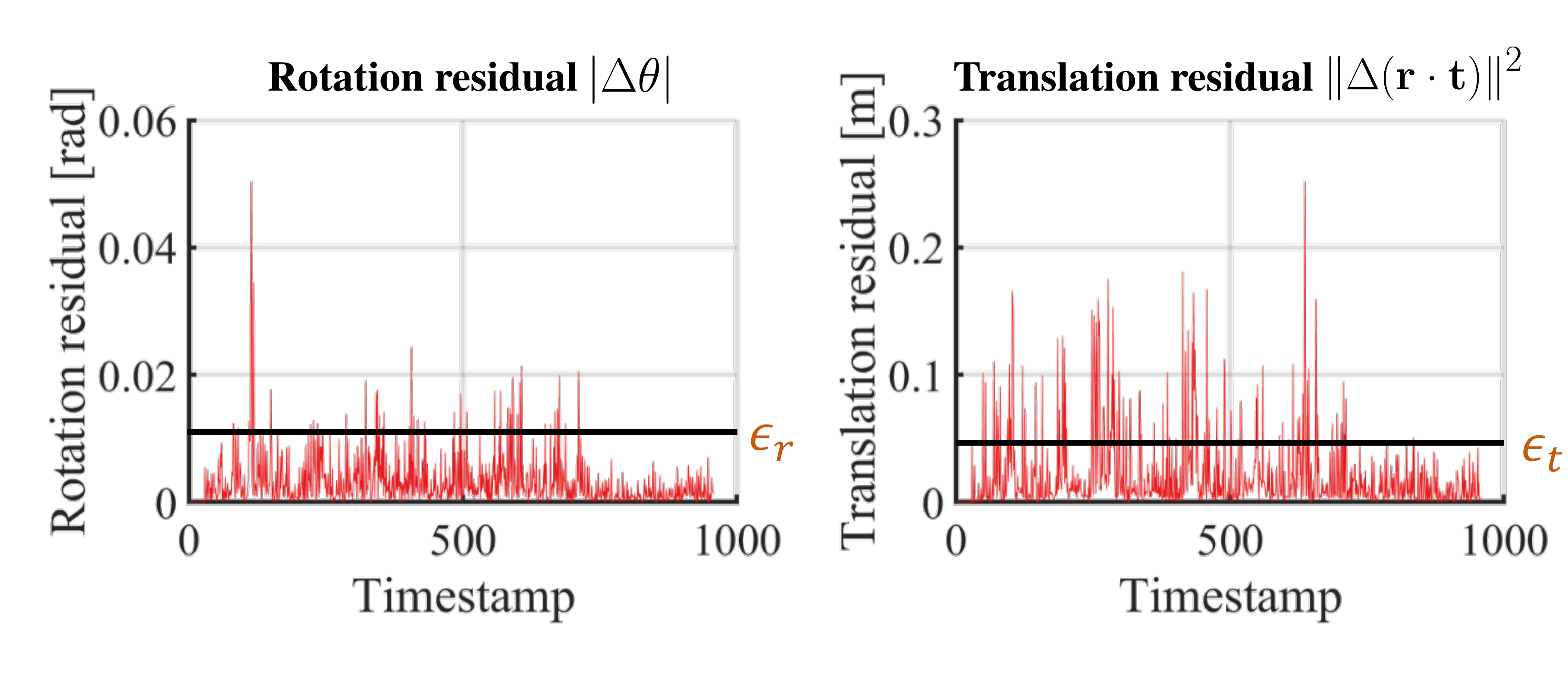}
			\caption{The screw motion residuals in rotation and translation of a set of estimated motions. These motions are visualized as the blue line in Fig. \ref{fig.rt} (left). $\epsilon_{r}, \epsilon_{r}$ are the thresholds to filter the outliers.}    
			\label{fig.screw_motion_example} 
			\vspace{-0.3cm}						
		\end{figure}

		We adopt the screw motion residuals to evaluate the performance of the previous motions estimation phase.
		Fig. \ref{fig.screw_motion_example} shows the screw motion residuals in a real-world example, where we find the estimated motions are very noisy. Hence, the outliers are filtered if both their rotation and translation residuals are larger than the thresholds: $\epsilon_{r}, \epsilon_{t}$. 
		The number of the filtered motions are denoted by $N$.

	\subsubsection{Pitch-roll rotation computation}	
		It is challenging to use rotation matrix to solve \eqref{equ.hand_eye_calibration_1} since the orthogonal constraint should be considered. For this reason, we employ the quaternion $(\mathbf{q}=[q_w, q_x, q_y, q_z]^{\top})$ following Hamilton notation to represent rotation. 
		\eqref{equ.hand_eye_calibration_1} can be thus rewritten by substituting $\mathbf{q}_{b}^{a} = (\mathbf{q}_{b}^{a})_{z}(\mathbf{q}_{b}^{a})_{yx}$ as below:
		\begin{align*}
			&\mathbf{q}_{a_{n}}^{a_{n-1}} \otimes (\mathbf{q}_{b}^{a})_{yx}
			=
			(\mathbf{q}_{b}^{a})_{yx} \otimes \mathbf{q}_{b_{n}}^{b_{n-1}}
			\\
			\Rightarrow \ 
			&\left[\mathbf{Q}_{1}(\mathbf{q}_{a_{n}}^{a_{n-1}}) 
			-
			\mathbf{Q}_{2}(\mathbf{q}_{b_{n}}^{b_{n-1}}) \right] \cdot (\mathbf{q}_{b}^{a})_{yx}
			\\
			\Rightarrow \
			&\mathbf{Q}_{n}^{n-1} \cdot (\mathbf{q}_{b}^{a})_{yx} = \mathbf{0},
			\numberthis		
			\label{equ.hand_eye_calibration_quaternion}			
		\end{align*}
		where 	
		\begin{align*}
			\mathbf{Q}_{1}(\mathbf{q}) 
			&= 
			\begin{bmatrix}
				q_w \mathbf{I}_{3} + \left[\mathbf{q}_{xyz}\right]_{\times} & \mathbf{q}_{xyz} \\
				-\mathbf{q}_{xyz}^{\top} & q_w
			\end{bmatrix} \\
			\mathbf{Q}_{2}(\mathbf{q}) 
			&= 
			\begin{bmatrix}
			q_w \mathbf{I}_{3} - \left[\mathbf{q}_{xyz}\right]_{\times}& \mathbf{q}_{xyz} \\
			-\mathbf{q}_{xyz}^{\top} & q_w
			\end{bmatrix}
			\numberthis
			\label{equ.hand_eye_calibration_quaternion_Q}				
		\end{align*}
		are matrix representations for left and right quaternion multiplication,  $\left[\mathbf{q}_{xyz}\right]_{\times}$ is the skew-symmetric matrix of $\mathbf{q}_{xyz}=
		[q_x,q_y,q_z]^{\top}$, and $\otimes$ is the quaternion multiplication operator. 

		With $N$ pairs of filtered rotations, we are able to formulate an over-constrained linear system as follows:
		\begin{align}
			\begin{bmatrix}
				\mathbf{Q}_{1}^{0} \\
				\mathbf{Q}_{2}^{1} \\
				\vdots \\
				\mathbf{Q}_{N}^{N-1} 
			\end{bmatrix}
			\cdot
			(\mathbf{q}_{b}^{a})_{yx}
			=
			\mathbf{Q}_{N}
			\cdot
			(\mathbf{q}_{b}^{a})_{yx} = \mathbf{0},
		\end{align}
		where $\mathbf{Q}_{N}$ is a $4N\times 4$ matrix. Using the singular value decomposition (SVD) $\mathbf{Q}_{N} = \mathbf{U}\mathbf{S}\mathbf{V}^{\top}$,  $(\mathbf{q}_{b}^{a})_{yx}$ is computed by the weighted sum of $\mathbf{v}_{3}$ and $\mathbf{v}_{4}$:
		\begin{align}
	    	(\mathbf{q}_{b}^{a})_{yx} &= \lambda_{1}\mathbf{v}_{3} + \lambda_{2}\mathbf{v}_{4},  
		\end{align}
		where $\mathbf{v}_{3}$ and $\mathbf{v}_{4}$ are the last two column vectors of $\mathbf{V}$, and $\lambda_{1}, \lambda_{2}$ are two scalars. Therefore, $(\mathbf{q}_{b}^{a})_{yx}$ can be obtained solving the following equations:
		\begin{align*}
			x_{(\mathbf{q}_{b}^{a})_{yx}}y_{(\mathbf{q}_{b}^{a})_{yx}} 
			&= 
			-z_{(\mathbf{q}_{b}^{a})_{yx}}w_{(\mathbf{q}_{b}^{a})_{yx}} \\
			\|(\mathbf{q}_{b}^{a})_{yx}\|
			&= 
			1,
			\numberthis		
			\label{equ.quaternion_constraint}				
		\end{align*}
		where $x_{\mathbf{q}}, y_{\mathbf{q}}, z_{\mathbf{q}}, w_{\mathbf{q}}$ are the elements of a quaternion.
	
	\subsubsection{Yaw rotation and translation computation}	
		Due to our planar motion assumption, the translation offset on $z\textendash$ axis is unobservable. Consequently, We set $t_z=0$ and rewrite \eqref{equ.hand_eye_calibration_2} by removing the third row as follows:
		\begin{align*}
			\mathbf{R}_{1}
			\begin{bmatrix}
				t_x \\
				t_y
			\end{bmatrix}
			-  
			\begin{bmatrix}
				\cos(\gamma) & -\sin(\gamma) \\
				\sin(\gamma) & \cos(\gamma)
			\end{bmatrix}
			\mathbf{t}_{1} = -\mathbf{t}_{2},
			\numberthis	
			\label{equ.hand_eye_calibration_2_2}
		\end{align*}
		where $t_x$ and $t_y$ are unknown translations along $x\textendash$ and $y\textendash$ axes, and $\gamma$ is the unknown rotation angle around $z\textendash$ axis. $\mathbf{R}_{1}$ is a $2\times 2$ upper-left submatrix of $(\mathbf{R}_{a_{n}}^{a_{n-1}}-\mathbf{I}_{3})$, $\mathbf{t}_1=[t_{11},t_{12}]^{\top}$ are the first two elements of $\mathbf{R}_{b}^{a}\mathbf{t}_{b_{n}}^{b_{n-1}}$, and $\mathbf{t}_{2}$ denote the first two elements of $\mathbf{t}_{a_{n}}^{a_{n-1}}$. We can rewrite \eqref{equ.hand_eye_calibration_2_3} as a matrix vector equation:
		\begin{align}
			\underbrace{		
			\begin{bmatrix}
				\mathbf{R}_{1} \ \mathbf{J}
			\end{bmatrix}}_{\mathbf{G}_{2\times 4}}
			\begin{bmatrix}
				t_{x} \\
				t_{y} \\
				-\cos(\gamma) \\
				-\sin(\gamma)
			\end{bmatrix},
			\label{equ.hand_eye_calibration_2_3}
		\end{align}
		where 
		$\mathbf{J} = 
		\begin{bmatrix}
		t_{11} & -t_{12} \\
		t_{12} & t_{11}
		\end{bmatrix}$.
		
		We can also construct a linear system from \eqref{equ.hand_eye_calibration_2_3} with the filtered motions: 
		\begin{align}
			\underbrace{
			\begin{bmatrix}
				\mathbf{G}_{1}^{0} \\
				\mathbf{G}_{2}^{1} \\
				\vdots \\
				\mathbf{G}_{N}^{N-1} \\
			\end{bmatrix}}_{\mathbf{A}_{2N\times 4}}
			\underbrace{
			\begin{bmatrix}
				t_{x} \\
				t_{y} \\
				-\cos(\gamma) \\
				-\sin(\gamma)
			\end{bmatrix}}_{\mathbf{x}_{4\times 1}}
			=
			-
			\underbrace{
			\begin{bmatrix}
				(\mathbf{t}_{2})_{1}^{0} \\
				(\mathbf{t}_{2})_{2}^{1} \\
				\vdots \\
				(\mathbf{t}_{2})_{N}^{N-1} 
			\end{bmatrix}}_{\mathbf{b}_{2N\times 1}},
			\label{equ.hand_eye_calibration_2_4}
		\end{align}
		where $\mathbf{x}$ is obtained by applying the least-squares approach.

\subsection{Refinement}
\label{sec.methodology_refinement}
	\begin{figure}[]
		\centering
		\includegraphics[width=0.3\textwidth]{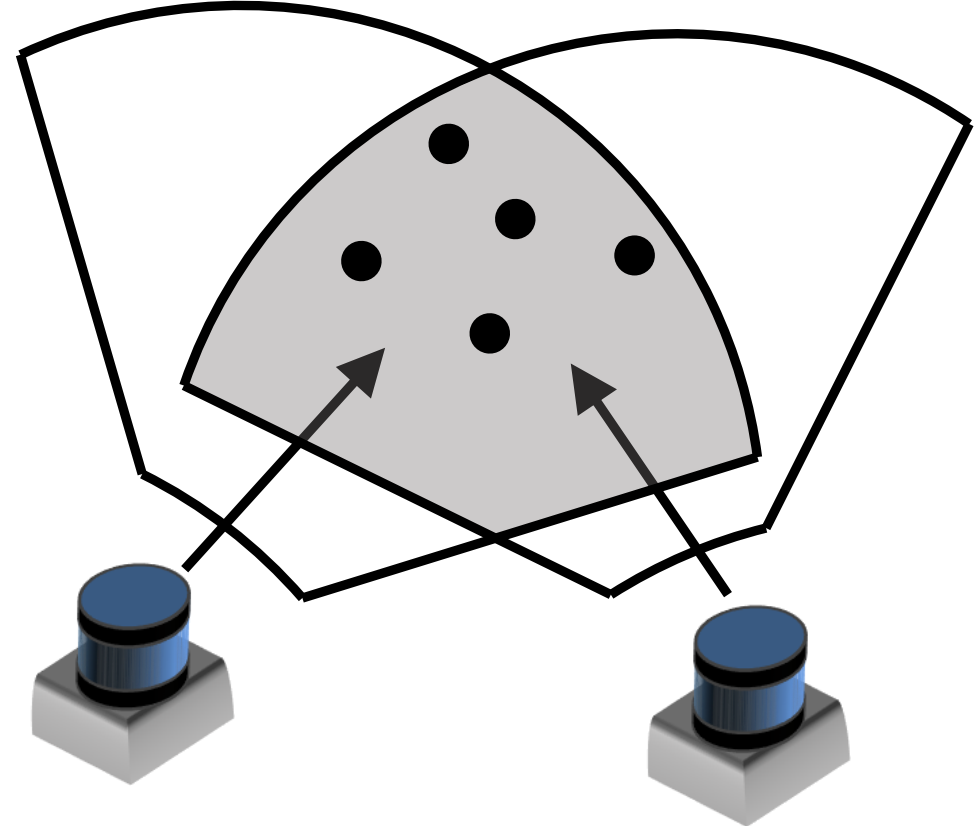}
		\caption{The FOV of each LiDAR is visualized partially, where the gray area is the overlapping region.}    
		\label{fig.overlapping_figure} 	
	\end{figure}  

	In this section, we combine the coarse initialization results with the sensor measurements to refine the extrinsic parameters. 
	Firstly, to recover the unknown $t_z$, the ground points are utilized. And then we estimate a set of transformations from $\{a\}$ to $\{b\}$ by registering all the available point clouds. To improve the registration accuracy, we apply an overlap filter to retain the points that lie in the overlapping regions. 
	
	\subsubsection{Ground planes alignment} 
	In the motion estimation phase, we have extracted $K$ pairs of ground points $\mathcal{G}^{a_k}, \mathcal{G}^{b_k}$ of the reference and target LiDARs. 
	Since the segmented point clouds are noisy, we implement the random sample consensus (RANSAC) plane fitting algorithm \cite{fischler1981random} to reject several outliers. 
	Denoting $\mathbf{c}_{k}^{a},\mathbf{c}_{k}^{b}$ as the centroids of $\mathcal{G}^{a_{k}}, \mathcal{G}^{b_{k}}$ after filtering, we use the mean value of $\big(\mathbf{c}_{k}^{a} - \mathbf{R}_{b}^{a}\mathbf{c}_{k}^{b}\big)_{z}$ at each timestamp to determine $t_z$.
	
	\subsubsection{Overlap Filter}
	\label{sec.overlap_estimation}
	
	The registration between these LiDARs is challenging since their overlapping field of view (FOV) is both limited and unknown. To tackle this issue, we propose an overlap filter to retains the points that lie within the counterpart LiDARs' FOV, as the gray area depicted in Fig. \ref{fig.overlapping_figure}.
	Denoting $\mathcal{P}^{a_{k}}, \mathcal{P}^{b_{k}}$ the point clouds captured by $a$ and $b$ at $k$ respectively, a point $\mathbf{p}\in\mathcal{P}^{b_{k}}$ can be transformed from $\{b\}$ to $\{a\}$ using the initial $\mathbf{T}_{b}^{a}$ such that:
	\begin{equation}
		\begin{split}
			\widetilde{\mathbf{p}} 
			=
			\mathbf{T}_{b}^{a}
			\mathbf{p}.
			\label{equ.point_cloud_projection}
		\end{split}
	\end{equation}
	
	Denoting $\mathcal{S}^{a_{k}} , \mathcal{S}^{b_k}$ the sets of points living in the volume of intersection between $\mathcal{P}^{a_{k}}, \mathcal{P}^{b_{k}}$. After transformation, we adopt the KD-Tree searching method \cite{bentley1975multidimensional} to construct them:
    \begin{equation}
    	\begin{split}
	    	\mathcal{S}^{a_{k}} 
	    	&= 
	    	\bigg\{\forall\mathbf{p}_{1}\in \mathcal{P}^{a_{k}}: d(\mathbf{p}_{1}, \widetilde{\mathbf{p}}_{2})<r, \exists \mathbf{p}_{2}\in\mathcal{P}^{b_{k}}\bigg\} \\
	    	\mathcal{S}^{b_k} 
	    	&= 
	    	\bigg\{\forall\mathbf{p}_{2}\in \mathcal{P}^{b_{k}}: d(\widetilde{\mathbf{p}}_{2}, \mathbf{p}_{1})<r, \exists \mathbf{p}_{1}\in \mathcal{P}^{a_{k}}\bigg\},
    	\end{split}
    \end{equation}
	where $\mathbf{p}_{1}$ and $\mathbf{p}_{2}$ represent an individual point from each point cloud, $d(\cdot, \cdot)$ is the Euclidean distance between two points, and $r$ is a threshold. We define the overlap factor as:
	\begin{equation}
		\begin{split}
			\Omega_{k} 
			= 
			\frac{|\mathcal{S}^{a_k}|}{|\mathcal{P}^{a_{k}}|}
			\cdot
			\frac{|\mathcal{S}^{b_k}|}{|\mathcal{P}^{b_{k}}|},
		\end{split}
	\end{equation}
	where $|\cdot|$ is the size of a set. 
	
	The point clouds $\mathcal{S}^{a_{k}} , \mathcal{S}^{b_k}$ with $\Omega_{k} > 0.8$ are selected as the inputs for the following registration step.

	\begin{figure}
		\vspace{0.25cm}
		\centering
		\includegraphics[width=0.43\textwidth]{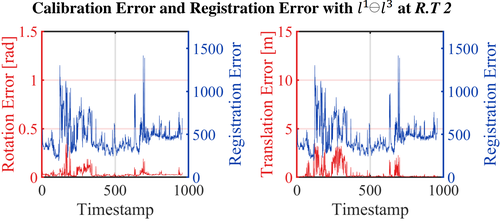}
		\caption{This figure depicts the curves calibration error and registration error with $l^{1}\ominus l^{3}$ at \textit{R.T 2}.}       
		\label{fig.registration} 	
		\vspace{-0.5cm}
	\end{figure}  	
		
	\subsubsection{Registration}	
    To calculate the relative transformation between two point clouds, the point-to-plane Iterative Closest Point (ICP) is used. After registering all the point cloud pairs captured at a different time, we obtain a set of transformations. For each result, its registration error and calibration error compared with ground truth are computed. An example of these errors over timestamp is depicted in
    Fig. \ref{fig.registration}, where we find that the curves of registration error and calibration error have a similar trends, especially the positions of the peak. For this reason, we only select a series of transformations with the minimum registration error as candidates and acquire the optimal refinement results by computing their mean values.

\section{Experiment}
\label{sec.experiment}

In this section, we divide the evaluation into two separate steps. 
Firstly, the initial calibration experiments are presented with simulated data and real sensor sets. 
Then we test the refined calibration on the real sensor data, and demonstrate that the initial results can be improved using appearance cues. 

\subsection{Implementation Details}
\label{subsec.implementation}

We use the Ceres Solver \cite{agarwal2012ceres} to solve the initialization problem and adopt the ICP library \cite{pomerleau2013comparing} to process point clouds. With empirically setting parameters: $\epsilon_{r} = 0.01, \epsilon_{t} = 0.01, r = 10$, our method can obtain promising results.
The success of motion-based calibration highly depends on the quality of the motions which a vehicle undergo. For this reason, we design several paths with different rotations and scales to test our proposed algorithm.
These paths include three simulated trajectories (\textit{S.T 1}-\textit{S.T 3}) and two real trajectories (\textit{R.T 1}-\textit{R.T 2}) on simulation and real sensor sets respectively. In the experiments, two platforms with different sensor setups are used for data collection.

\subsubsection{Simulation}
The simulated car platform\footnote{\url{https://www.osrfoundation.org/simulated-car-demo/}}
is a well known publicly available software. 
For testing, we manually mount two sensors at different positions on the platform, as shown in Fig. \ref{fig.simulation_setup} (left). The rotation offset between them is approximately $[0,3.14,1.57]$ rads in roll, pitch, and yaw respectively. The corresponding translation are approximately $[-2.5,1.5,0]$ meters along $x\textendash, y\textendash,$ and $z\textendash$ axes respectively. 
We can thus acquire the positions of these sensors in the form of ground truth at $5$ Hz. The refinement is not tested in simulation because this platform does not provide stable point clouds without time distortion. 

\subsubsection{Real Sensor}
While our approach is not limited to a particular number of target sensors, we are specifically interested in calibrating between the reference LiDAR and two target LiDARs based on our platform.
As shown in Fig. \ref{fig.platform}, 
three 16-beam RS-LiDARs\footnote{\url{https://www.robosense.ai/rslidar/rs-lidar-16}} are rigidly mounted on the vehicle. 
The setup of this multi-LiDAR system has significant transformation among sensors. 
Especially, $l^3$ is mounted with approximately $180$ degree rotation offset in yaw.
In later sections, we use $l^{1} \ominus l^{i}$ to represent the configuration between $l^{1}$ and $l^{i}$.
Since we do not know the precise extrinsic parameters of the multi-LiDAR system, we use the parameters (shown in Table \ref{tab.real_sensor_setup}) provided by the manufacturer to evaluate our proposed algorithm.

With respect to the accuracy calculation, 
the error in rotation is measured according to the angle difference between the ground truth $\mathbf{R}_{\text{gt}}$ and the resulting rotation $\mathbf{R}_{\text{resulting}}$, which is calculated as $e_{r} = \|\log(\mathbf{R}_{\text{gt}}^{}\mathbf{R}_{\text{resulting}}^{-1})^{\vee}\|_{2}$. 
Similarly, the difference in translation is computed using vector subtraction as $e_{t} = \|\mathbf{t}_{\text{gt}}^{}-\mathbf{t}_{\text{resulting}}^{}\|_{2}$. The translation error on $z\textendash$ axis will not be counted of the initialization results because of the planar movement assumption.

\subsection{Performance of Initialization}
\label{subsec.initialization}

We take the modified Kabsch algorithm \cite{kabsch1978discussion,taylor2015motion} that operates at matrix representation for comparison.

\subsubsection{Simulation}
\begin{figure}
	\centering
	\includegraphics[width=0.45\textwidth]{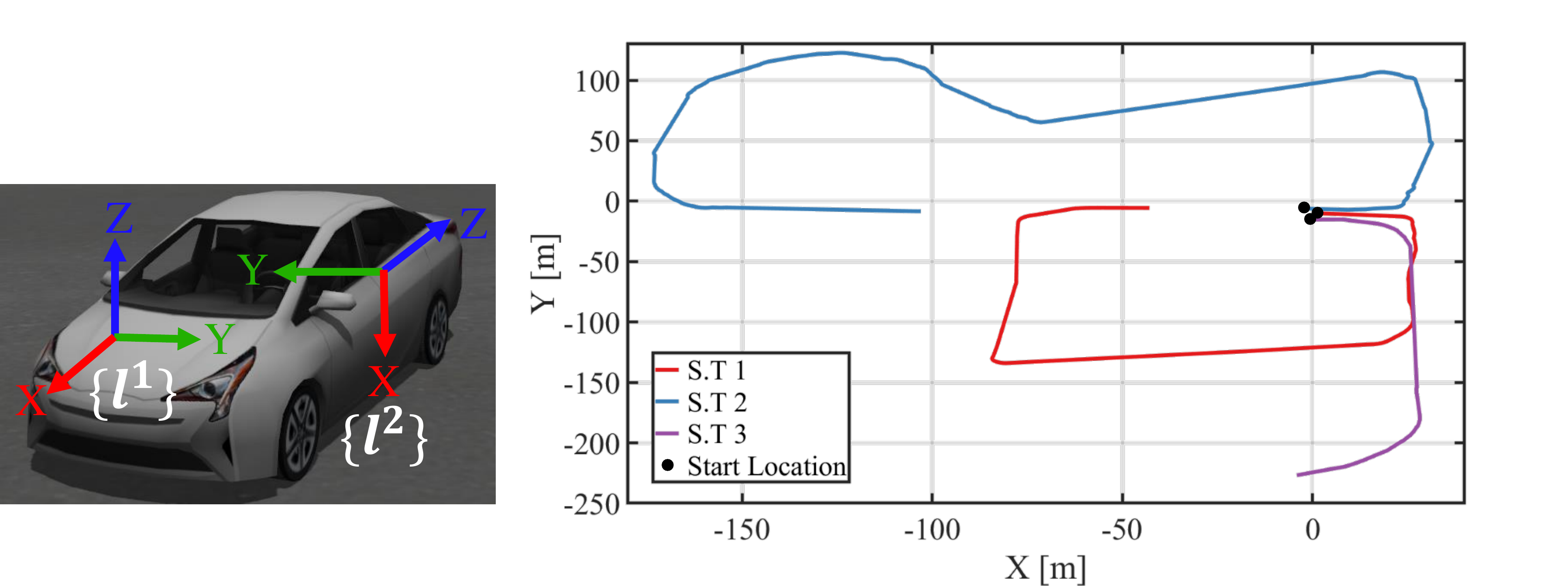}       
	\caption{(Left) The simulated platform and (right) the three trajectories which the platform follows.The rotation offset is about $[0,3.14,1.57]$ rads in roll, pitch, and yaw respectively. The corresponding translation are about $[-2.5,1.5,0]$ meters along $x\textendash, y\textendash,$ and $z\textendash$ axes respectively.}    
	\label{fig.simulation_setup} 	
\end{figure} 

\begin{table}[]
	\centering
	\caption{The initial calibration results with simulated data.}
	\renewcommand\arraystretch{0.9}
	\renewcommand\tabcolsep{5pt}	
	\begin{tabular}{cccccccc}
		\hline
		\toprule
		\multicolumn{3}{c}{\multirow{2}{*}{$\sigma^{2}$}} & \multirow{2}{*}{Trajectory} & \multicolumn{2}{c}{Rotation Error {[}rad{]}} & \multicolumn{2}{l}{Translation Error {[}m{]}} \\ \cline{5-8}
		\multicolumn{3}{c}{}                                              &                        & Kabsch                & Proposed             & Kabsch                 & Proposed             \\ 
		\hline		 
		\multicolumn{3}{c}{\multirow{3}{*}{$\sigma_1^{2}$}}                       & \textit{S.T 1}                      & 0.10               & \textbf{0.01}             & \textbf{0.22}               & 0.28             \\
		
		\multicolumn{3}{c}{}                                              & \textit{S.T 2}                      & 0.80              & \textbf{0.00}             & 0.66               & \textbf{0.28}             \\
		
		\multicolumn{3}{c}{}                                              & \textit{S.T 3}                      & 0.08              & \textbf{0.01}             & 0.80               & \textbf{0.48}             \\
		
		\hline		
		\multicolumn{3}{c}{\multirow{3}{*}{$\sigma_2^{2}$}}                        & \textit{S.T 1}                      &1.37              &\textbf{0.07}             & 2.13               & \textbf{1.25}             \\
		
		\multicolumn{3}{c}{}                                              & \textit{S.T 2}                     & 0.94              & \textbf{0.04}             & 1.80               & \textbf{1.20}             \\
		
		\multicolumn{3}{c}{}                                              & \textit{S.T 3}                      & 1.17              & \textbf{0.02}             & 1.66              & \textbf{1.44}            \\
		
		\hline
		\toprule		
		\label{tab.initialization_simulation_result}    		
	\end{tabular}	
	\vspace{-0.6cm}	
\end{table}	

The simulated trajectories \textit{(S.T 1-S.T 3}) are visualized in Fig. \ref{fig.simulation_setup} (right), where the third one is considered as the most challenging one since it has fewer rotations.
To verify the performance of our proposed algorithm, the sensor's motions are added with zero-mean Gaussian noise $\mathbf{n}\sim\mathcal{N}(\mathbf{0}, \sigma^{2})$.
$\sigma^{2}$ is set to two values: $\sigma_1^{2}=0.0001$ and $\sigma_2^{2}=0.001$ for evaluation. 
For each value, all simulated motions are tested.
The calibration results are shown in Table \ref{tab.initialization_simulation_result}. 
The proposed algorithm can successfully initialize the rotation offset with low error, and outperform the Kabsch algorithm in most of cases.
We also note that the translation error with $\sigma_{2}^{2}$ is larger than $1 \text{m}$, meaning that our initialization approach is not robust in noisy data.
But the usage of appearance-based refinement can solve this problem.

\subsubsection{Real Sensor}
\begin{figure}
	\vspace{0.2cm}  	
	\centering
	\includegraphics[width=0.35\textwidth]{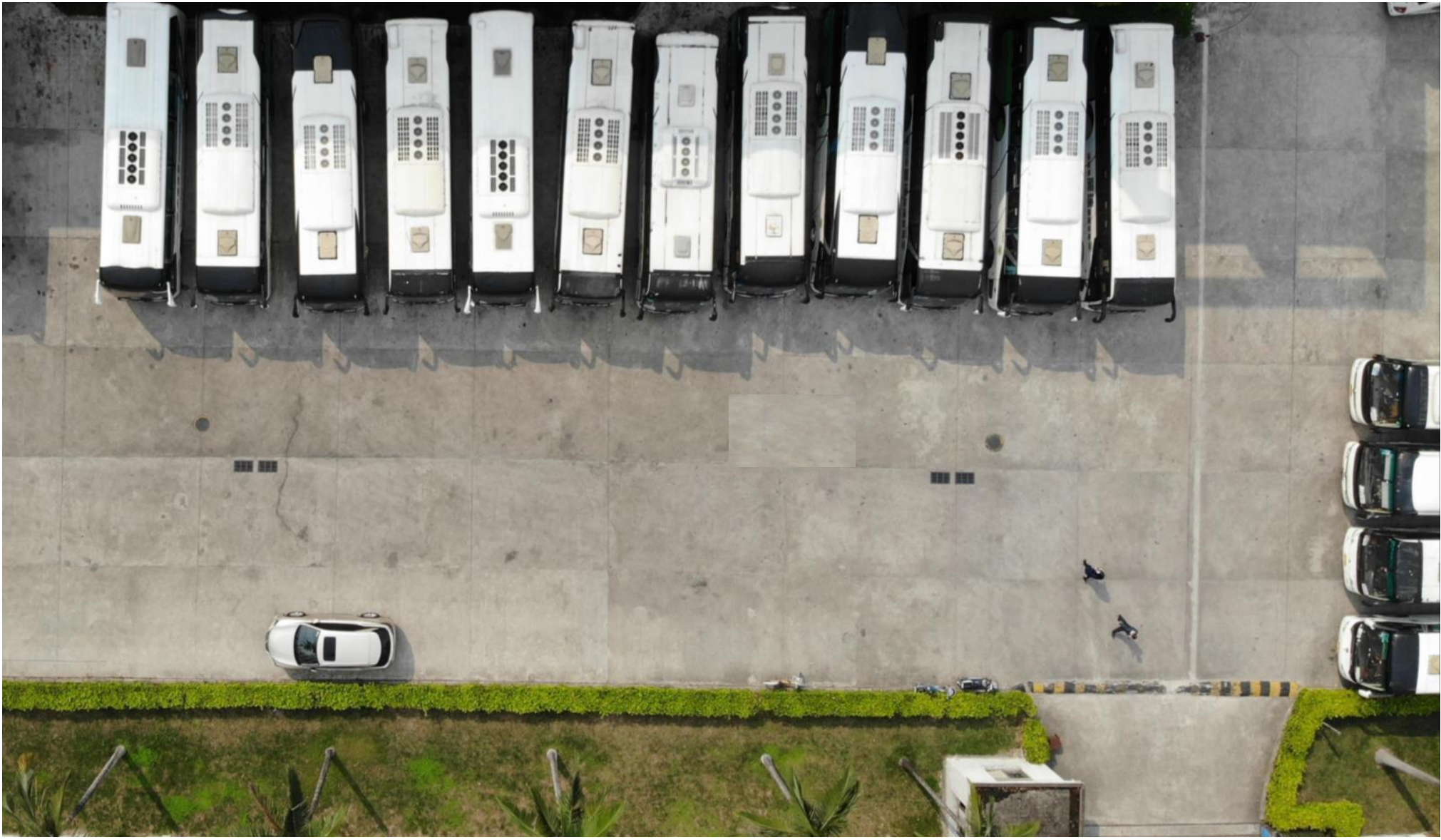}
	\caption{The testing environment for our calibration experiments.}       
	\label{fig.real_sensor_scene} 	
\end{figure}  

\begin{figure}
	\centering
	\includegraphics[width=0.4\textwidth]{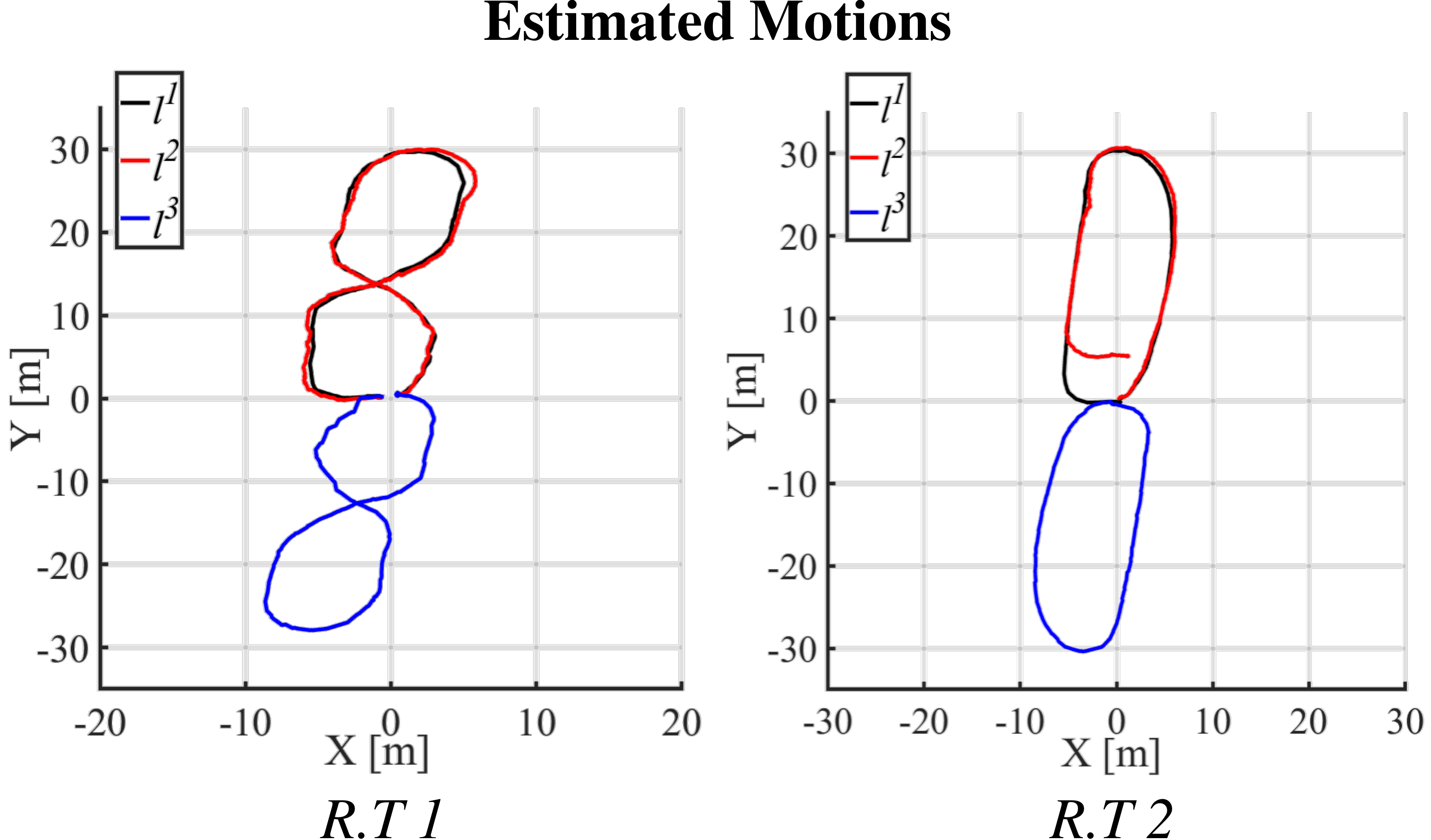}
	\caption{The black, red, and blue lines indicate the estimated motions of $l_1, l_2, l_3$ respectively at \textit{R.T 1-R.T 2}. }       
	\label{fig.rt} 	
	\vspace{-0.5cm}	
\end{figure}

We carry out a real sensor experiment to validate the proposed method. Two trajectories (\textit{R.T 1-R.T 2}) are designed in an urban environment (shown in \ref{fig.real_sensor_scene}), and we drive the vehicle to follow these trajectories. After estimating the individual motions of each LiDAR, we can use these results to initial the calibration. The estimated motions are depicted in Fig. \ref{fig.rt}. We observe that the calculated trajectory of $l^{2}$ at \textit{R.T 2} drifts.
The initialization results are presented in Table \ref{tab.initialization_result}. Our method performs well in recovering the rotation offset $(<0.15 \text{rad})$, but fail in calculating the translation offset $(>0.5 \text{m})$ in all cases. 
With the above results in simulation and real-world environments, we can conclude that the initialization phase can provide coarse estimates to the extrinsic parameters. For precise results, an additional refinement step is required.

\subsection{Performance of Refinement}
\label{subsec.refinement}

\begin{figure}[t]
	\centering
	\includegraphics[width=0.5\textwidth]{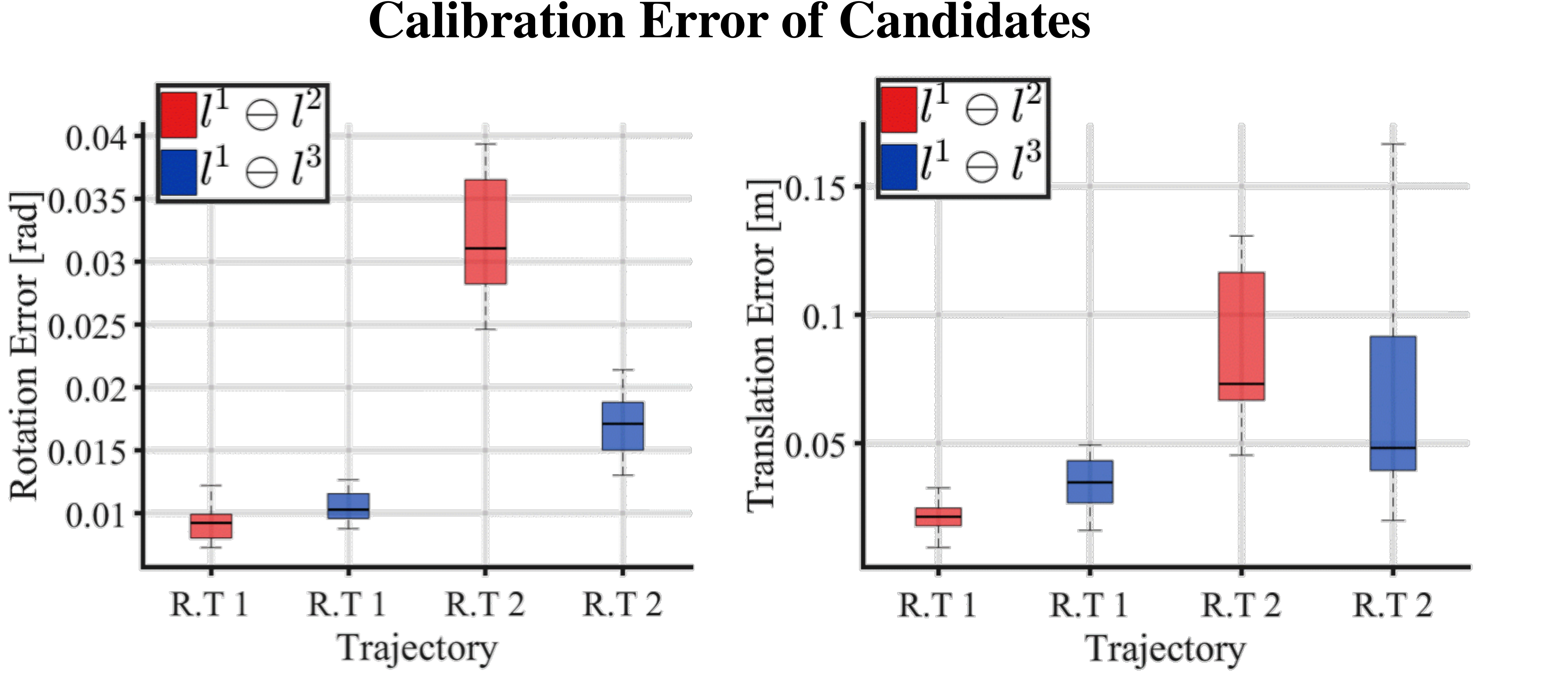}
	\caption{The calibration errors in each case of the candidate. The means of them are small at \textit{R.T 1}, but large at \textit{R.T 2}.}       
	\label{fig.refinement_selected_error} 	
\end{figure}  

\begin{table}[t]
	\centering
	\caption{The calibration ground truth of our multi-lidar system.}	
	\renewcommand\arraystretch{0.9}
	\renewcommand\tabcolsep{5pt}	
	\begin{tabular}{ccccccc}
		\hline
		\toprule
		\multicolumn{1}{c}{\multirow{2}{*}{Conf.}} & \multicolumn{3}{c}{Rotation {[}rad{]}}                                               & \multicolumn{3}{c}{Translation {[}m{]}}    \\ \cline{2-7}
		\multicolumn{1}{c}{}                       & x                         & y                         & z                          & x                          & y                         & z                          \\ \hline
		$l^{1}\ominus l^{2}$         & 0.01                           & 0.08                      & 0.03                       & 0.42                       & 0.00                      & -1.26                      \\ 
		$l^{1}\ominus l^{3}$                             & \multicolumn{1}{c}{-0.02} & \multicolumn{1}{c}{0.01} & \multicolumn{1}{c}{-3.11} & \multicolumn{1}{c}{-2.11} & \multicolumn{1}{c}{0.06} & \multicolumn{1}{c}{-1.18} \\ 
		\hline
		\toprule
	\end{tabular}	
	\vspace{-0.2cm}	
	\label{tab.real_sensor_setup} 	
\end{table}

\begin{table}[!htp]
	\centering
	\caption{The initialization results.}	
	\renewcommand\arraystretch{0.9}
	\renewcommand\tabcolsep{5pt}	
	\begin{tabular}{ccccccccccc}
		\hline
		\toprule
		\multicolumn{2}{c}{\multirow{2}{*}{Conf.}}                & \multirow{2}{*}{Traj.} & \multicolumn{4}{c}{Rotation Error {[}rad{]}}                    & \multicolumn{4}{c}{Translation Error {[}m{]}}                          \\ \cline{4-11} 
		\multicolumn{2}{c}{}                                      &                        & \multicolumn{2}{c}{Kabsch} & \multicolumn{2}{c}{Proposed}      & \multicolumn{2}{c}{Kabsch}        & \multicolumn{2}{c}{Proposed}      \\ \hline \toprule
		\multicolumn{2}{c}{\multirow{2}{*}{$l^{1}\ominus l^{2}$}} & \textit{R.T 1}                 & \multicolumn{2}{c}{0.94}   & \multicolumn{2}{c}{\textbf{0.06}} & \multicolumn{2}{c}{1.60}          & \multicolumn{2}{c}{\textbf{0.47}} \\ 
		\multicolumn{2}{c}{}                                      & \textit{R.T 2}                  & \multicolumn{2}{c}{0.73}   & \multicolumn{2}{c}{\textbf{0.03}} & \multicolumn{2}{c}{4.42}          & \multicolumn{2}{c}{\textbf{1.95}} \\ \hline
		\multicolumn{2}{c}{\multirow{2}{*}{$l^{1}\ominus l^{3}$}} & \textit{R.T 1}                  & \multicolumn{2}{c}{1.29}   & \multicolumn{2}{c}{\textbf{0.14}} & \multicolumn{2}{c}{\textbf{1.23}} & \multicolumn{2}{c}{1.26}          \\ 
		\multicolumn{2}{c}{}                                      & \textit{R.T 2}                  & \multicolumn{2}{c}{0.60}   & \multicolumn{2}{c}{\textbf{0.08}} & \multicolumn{2}{c}{\textbf{1.36}} & \multicolumn{2}{c}{2.04}          \\ 
		\hline
		\toprule
	\end{tabular}
	\vspace{-0.2cm}	
	\label{tab.initialization_result} 	
\end{table}

\begin{table}[!htp]
	\centering
	\caption{The refinement results.}
	\renewcommand\arraystretch{0.9}
	\renewcommand\tabcolsep{5pt}					
	\begin{tabular}{ccccccc}
		\hline
		\toprule
		\multicolumn{2}{c}{Conf.}          & $l^{1}\ominus l^{2}$ & $l^{1}\ominus l^{2}$ & $l^{1}\ominus l^{3}$ & \multicolumn{2}{c}{$l^{1}\ominus l^{3}$} \\ 
		\multicolumn{2}{c}{Traj.}             & \textit{R.T 1}           & \textit{R.T 2}           & \textit{R.T 1}         & \multicolumn{2}{c}{\textit{R.T 2}} \\ \hline \toprule
		\multirow{4}{*}{Rotation {[}rad{]}}  & x     & 0.01            & 0.01            & -0.02         & \multicolumn{2}{c}{-0.02} \\  
		& y     & 0.08            & 0.07            & 0.02          & \multicolumn{2}{c}{0.00}  \\ 
		& z     & 0.04            & 0.04            & -3.14         & \multicolumn{2}{c}{-3.13} \\ 
		& \textbf{error} & 0.01            & 0.03            & 0.01         & \multicolumn{2}{c}{0.02}  \\ 
		\hline
		\multirow{4}{*}{Translation {[}m{]}} & x     & 0.43            & 0.46            & -2.13         & \multicolumn{2}{c}{-2.07} \\  
		& y     & 0.00            & -0.01           & 0.09          & \multicolumn{2}{c}{0.08}  \\  
		& z     & -1.26           & -1.27           & -1.18         & \multicolumn{2}{c}{-1.20} \\  
		& \textbf{error} & 0.01            & 0.08            & 0.03          & \multicolumn{2}{c}{0.04}  \\ 
		\hline
		\toprule
	\end{tabular}
	\label{tab.refinement_results}
	\vspace{-0.3cm}
\end{table}

In our experiments, we select $10$ resulting transformations as the candidates for the optimal refinement results.
We plot their calibration errors in each case in Fig. \ref{fig.refinement_selected_error}.
The detailed calibration results are shown in Table \ref{tab.refinement_results}, where all the rotation and translation errors are less than 0.04 [rad] and 0.1 [m] respectively. Compared with the result in Table \ref{tab.initialization_result}, the refinement phase can improve the estimated parameters.

\subsection{Discussion}
\label{sec.discussion}

Since our proposed method achieves accurate calibration results, but its performance will be influenced by the below conditions. Firstly, the initialization relies on the accuracy of estimated motions, especially when the translation offset is imprecise. Furthermore, we assume that the LiDARs' view should have overlapping regions. Finally, the registration between point clouds will fail in several feature-less scenes.

\section{Conclusion and Future Work} 
\label{sec.conclusion}

In this paper, we presented a novel method for automatically calibrating of a multi-LiDAR system without any extra sensors, calibration target, or prior knowledge about surroundings. 
Our approach makes use of the complementary strengths of the motion-based and appearance-based calibration methods, which consists of three phases. The individual motions of each LiDAR are estimated by a LiDAR odometry algorithm. The calculated motions are then utilized to initialize the extrinsic parameters. Finally the results are refined by exploiting appearance cues in sensors' overlap.
The performance of our method is demonstrated through a series of simulated and real-world experiments with reliable and accurate calibration results. There are several possible extensions to this work, (1) online multi-LiDAR calibration, (2) calibration without the overlapping requirement, (3) applications based on a multi-LiDAR system.

\clearpage
\balance
\bibliographystyle{IEEEtran}
\bibliography{reference}{}

\end{document}